\documentclass[runningheads]{llncs}
\usepackage[T1]{fontenc}
\usepackage{graphicx}
\usepackage{subcaption}
\usepackage{tabularx}
\usepackage{multirow}
\usepackage{bbm}
\usepackage{amsmath}
\usepackage{hyperref}
\newcolumntype{Y}{>{\centering\arraybackslash}X}

\begin{document}

\title{Improving Early Sepsis Onset Prediction Through Federated Learning}

\author{Christoph Düsing\orcidID{0000-0002-7817-9448} \and
Philipp Cimiano\orcidID{000-0002-4771-441X}}
\authorrunning{C. Düsing \& P. Cimiano}
\institute{CITEC, Bielefeld University, Bielefeld, Germany \\ 
\email{cduesing}@techfak.uni-bielefeld.de}

\maketitle 

\begin{abstract}
Early and accurate prediction of sepsis onset remains a major challenge in intensive care, where timely detection and subsequent intervention can significantly improve patient outcomes. While machine learning models have shown promise in this domain, their success is often limited by the amount and diversity of training data available to individual hospitals and \textit{Intensive Care Units} (ICUs). \textit{Federated Learning} (FL) addresses this issue by enabling collaborative model training across institutions without requiring data sharing, thus preserving patient privacy. In this work, we propose a federated, attention-enhanced Long Short-Term Memory model for sepsis onset prediction, trained on multi-centric ICU data. Unlike existing approaches that rely on fixed prediction windows, our model supports variable prediction horizons, enabling both short- and long-term forecasting in a single unified model. During analysis, we put particular emphasis on the improvements through our approach in terms of early sepsis detection, i.e., predictions with large prediction windows by conducting an in-depth temporal analysis. 
Our results prove that using FL does not merely improve overall prediction performance (with performance approaching that of a centralized model), but is particularly beneficial for early sepsis onset prediction. Finally, we show that our choice of employing a variable prediction window rather than a fixed window does not hurt performance significantly but reduces computational, communicational, and organizational overhead. 

\keywords{Federated Learning \and Sepsis Onset Prediction \and Recurrent Neural Network.}
\end{abstract}

\section{Introduction}

In recent years, \textit{Clinical Decision Support Systems} (CDSSs) have seen increasing demand, particularly due to the growing adoption of \textit{Electronic Health Records} (EHRs) and the need for tools to assist in timely and adequate therapy. In this context, various \textit{Machine Learning} (ML) techniques have facilitated remarkable improvements in CDSS quality, with applications ranging from diagnosis to prognosis and treatment recommendation \cite{tonekaboni2019clinicians}. These systems are particularly valuable in intensive care settings, where critically ill patients benefit the most from real-time recommendations. Among the most critical and time-sensitive conditions encountered in \textit{Intensive Care Units} (ICUs) is sepsis, a life-threatening response to an infection that requires timely identification and intervention \cite{singer2016third}. 

While the improvements in the quality of CDSSs through ML -- and \textit{Deep Learning} (DL) in particular -- are impressive, they come at the cost of requiring large and diverse amounts of EHRs to learn from. Unfortunately, in the context of sepsis, it has been shown that average U.S. hospitals do not possess sufficient volumes of EHRs to reliably train such models \cite{dusing2024integrating,dusing2023federated}. This raises concerns regarding the ability of smaller hospitals to benefit from these technologies, especially when data centralization is restricted due to privacy concerns.

To address this limitation and avoid direct data sharing among institutions, \textit{Federated Learning} (FL) has emerged as a promising paradigm \cite{fang2022dp}. FL enables multiple institutions to collaboratively train a shared model while keeping patient data private \cite{mcmahan2017communication,kairouz2021advances}. This makes FL particularly well-suited for the healthcare domain, where data privacy regulations such as the \textit{Health Insurance Portability and Accountability Act} (HIPAA) and the \textit{General Data Protection Regulation} (GDPR) strictly limit data sharing across institutional boundaries.

FL has already demonstrated success in various medical applications beyond sepsis, including kidney injury prediction \cite{rajendran2023data}, ICU mortality prediction \cite{mondrejevski2023predicting}, and drug discovery \cite{oldenhof2023industry}. These studies demonstrate that FL can be effectively applied in settings where data privacy is a major concern. In the specific context of sepsis, prior work includes the deployment of FL for therapy recommendation \cite{dusing2023federated,dusing2024integrating} and for sepsis onset prediction with fixed prediction windows sizes \cite{pan2024adaptive}.

While the existing body of work shows that the application of FL in the medical domain is not novel per se and has demonstrated promising results -- particularly in sepsis prediction and therapy support -- we argue that our study advances beyond these existing approaches. Specifically, the novelty and contributions of our work are as follows:
\begin{enumerate}
    \item \textbf{Variable Prediction Windows:} Unlike most existing models that rely on fixed prediction horizons (e.g., \cite{mondrejevski2023predicting,pan2024adaptive}), we propose an attention-enhanced \textit{Long Short-Term Memory} (LSTM) model, which supports a variable prediction window (25h to 1h predictions). This enables both short-term predictions of sepsis onset at the time of admission as well as long-term forecasts, increasing the model's flexibility and practical utility while reducing the overhead associated with maintaining multiple specialized models \cite{goh2021artificial}.
    \item \textbf{Improved Early Prediction Through FL:} Although early identification is critical, prior research has largely overlooked the specific challenge of early sepsis prediction \cite{zhou2024early}. To this end, we provide an in-depth analysis demonstrating that FL does not merely improve prediction performance uniformly across time but is particularly effective at enhancing early sepsis detection. Our results show that FL provides more significant improvements as the prediction horizon increases, thus allowing for earlier interventions and potentially better clinical outcomes.
    \item \textbf{Focus on Multi-Centric Settings:} Unlike previous approaches that often conduct mono-centric analyses, our study highlights the utility of FL in a multi-ICU setting. This is especially important in clinical research, where multi-centric studies are widely considered the gold standard for achieving reliable and generalizable results \cite{zhou2024early}.
\end{enumerate}

\section{Related Work}

Sepsis onset prediction has been previously studied to some extent due to its critical importance in improving patient outcomes, especially in intensive care settings \cite{moor2021early}. Such early warning systems can help physicians identify deteriorating patients and enable more timely interventions, potentially reducing mortality.

Initial efforts in sepsis prediction often employed classical ML models such as \textit{RandomForests} \cite{breiman2001random}. For instance, Zhou et al.\ \cite{zhou2024early} proposed a model that utilizes this technique alongside unbalanced data processing to predict early sepsis in a clinical context. Similarly, Rajendran et al.\ \cite{rajendran2023data} used a \textit{Multi-Layer Perceptron}, while Pan et al.\ \cite{pan2024adaptive} developed an ensemble-based approach for sepsis prediction. While these models can effectively capture patterns in tabular patient data, they are less suitable for modeling temporal dependencies inherent in most EHRs. This limitation is especially relevant in sepsis care, where recognizing dynamic and temporal changes -- such as trends in vital signs or lab values -- is essential for timely intervention. For example, the widely adopted \textit{Sepsis-3} definition \cite{singer2016third} considers sepsis as a rise of 2 or more points in the \textit{Sequential Organ Failure Assessment} (SOFA) score, regardless of its absolute value, proving the importance of tracking trends over time.

To better capture the temporal nature of patient data, more recent approaches have leveraged \textit{Recurrent Neural Networks} (RNNs), particularly LSTM networks \cite{hochreiter1997long}. Svenson et al.\ \cite{svenson2020sepsis}, for example, deployed LSTMs to predict deteriorating conditions in sepsis patients. Similarly, Scherpf et al.\ \cite{scherpf2019predicting} and Zhang et al.\ \cite{zhang2021interpretable} applied RNN-based models to forecast sepsis onset directly from multivariate time-series data. Other works have expanded on this with more advanced architectures such an attention-based models for predicting sepsis mortality \cite{mondrejevski2024masicu}. 

More recently, such time-series-based models have been extended to the FL setting. As discussed earlier, FL enables multiple institutions to collaboratively train a shared model without exchanging raw patient data, thereby preserving privacy in compliance with regulations like HIPAA and GDPR \cite{mcmahan2017communication}. Technically, FL involves multiple rounds of training, where each round consists of: (1) distributing a global model to all participating clients; (2) clients locally updating the model using their private data; and (3) aggregating these local updates on a central server to produce a new global model \cite{mcmahan2017communication,kairouz2021advances}. Following this paradigm, Mondrejevski et al. introduced \textit{FLICU}, a federated LSTM-based model designed for ICU mortality prediction across multiple hospitals without data centralization \cite{mondrejevski2022flicu}. Building on this foundation, they later applied a similar framework to predict sepsis onset \cite{mondrejevski2023predicting}. However, both models employ a fixed \textit{prediction window} -- i.e., they are trained to forecast sepsis a pre-determined number of hours in advance. To compare performance across different prediction windows, some studies train multiple separate models with different prediction windows (e.g., 3h, 6h, or 12h) \cite{mondrejevski2023predicting,goh2021artificial,rosnati2021mgp,scherpf2019predicting}. The general consensus across these studies is that predictive accuracy tends to decline as the prediction window increases.

Despite their contributions, two key limitations remain across the majority of these studies. First, they rely exclusively on fixed prediction windows, limiting the flexibility and practical deployment of models. Some attempt to address this by training multiple separate models for different prediction horizons, but this adds substantial overhead. Second, although many works observe that performance declines for longer prediction windows, they do not conduct in-depth analyses of whether FL or alternative strategies can mitigate this degradation. In particular, no comprehensive evaluation has been provided to assess if FL contributes more effectively to early detection compared to local models.

\section{Methodology}

Our approach aims to enhance early sepsis prediction using FL to collaboratively train an LSTM model. This model processes sequences of patient features to predict whether a patient is likely to develop sepsis during their ICU stay -- without requiring data centralization. 
Beyond maintaining privacy, FL reduces regulatory burdens compared to alternatives such as shared data spaces, which require exchanging sensitive patient data across institutions. 
We begin by describing the dataset acquisition steps for this study. We then introduce the prediction model and detail how it is trained across multiple ICUs in a federated manner. Finally, we outline the evaluation strategy used to assess the model performance.

\subsection{Data Acquisition}

This study is based on the publicly available \textit{MIMIC-IV} dataset \cite{mimic}, which contains de-identified EHRs of ICU patients admitted to the \textit{Beth Israel Deaconess Medical Center} and was chosen for its public availability, which facilitates reproducibility and broader adoption. The dataset includes admissions from seven ICUs: \textit{Medical ICU} (MICU), \textit{Medical/Surgical ICU} (MICU/SICU), \textit{Surgical ICU} (SICU), \textit{Trauma SICU} (TSICU), \textit{Coronary Care Unit} (CCU), \textit{Cardiac Vascular ICU} (CVICU), and \textit{Neuro SICU} (NSICU).

\noindent{\textbf{Patient Inclusion.}} To ensure data quality and consistency, we applied the following patient inclusion criteria, inspired by prior work \cite{mondrejevski2022flicu}:
\begin{itemize}
    \item Only the first ICU stay per patient is considered.
    \item Patients with ICU stays shorter than 30 hours are excluded to ensure a sufficient observation window.
    \item Patients admitted to the \textit{Neonatal} or \textit{Pediatric} ICUs are excluded.
\end{itemize}

After applying these criteria, the final dataset comprises 28,610 patients. Each patient is then assigned to the ICU they were initially admitted to, resulting in a naturally partitioned dataset for FL, where each ICU resembles a client. Among other things, the number of patients per ICU are outlined in Table~\ref{table:baelines}.

\noindent{\textbf{Data Pre-processing.}} The selection of features was guided by prior studies on sepsis onset prediction and therapy recommendation \cite{dusing2023federated,dusing2024integrating,mondrejevski2023predicting}. In total, we selected 26 clinically relevant features, grouped into four categories: \textit{general}, \textit{vital signs}, \textit{diagnoses}, and \textit{therapies}. These are detailed in Table~\ref{table:features}.

\begin{table}[t]
    \begin{center}
        \caption{Feature Selection for Sepsis Onset Prediction}\label{table:features}
        \scalebox{0.95}{
        \begin{tabularx}{\textwidth}{ c || Y} 
            Category & Features\\
            \hline \hline
            General Features & Gender, Ethnicity, Age, Height, Weight \\
            \hline
            Vital Features & Platelet, Leukocytes, PO2, FiO2, Lactate, Creatinine, Bilirubin, Glasgow-Coma-Scale, C-reactive protein, Diastolic Pressure, Systolic Pressure, Mean Blood Pressure, Respiratory Rate, Temperature, SpO2, Urine Output, Glucose, Heart Rate \\
            \hline
            Diagnosis Features & Diabetes, SOFA\\
            \hline
            Therapy Features & Mechanical Ventilation\\
        \end{tabularx}
        }
    \end{center}
\end{table}

Then, all features are aggregated into 1-hour time windows following ICU admission to balance temporal resolution with data sparsity. If multiple values for a feature occur within the same hour, the most recent measurement is retained. Missing values within a patient’s time-series are imputed via linear interpolation, while features entirely missing for a patient are replaced with the global mean to provide a neutral value when no patient-specific information is available.

We collect data for the first 30 hours of each patient’s ICU stay. This time frame is motivated by previous findings suggesting that clinical patterns up to 20-30 hours before sepsis onset are relevant to its prediction \cite{zhang2021interpretable}. This results in multivariate time-series with 30 hourly measurements for each of the 26 features.

Subsequently, sepsis labels are assigned following the \textit{Sepsis-3} definition \cite{singer2016third}. Accordingly, each patient is labeled as positive if they developed sepsis within the first 30 hours of ICU admission.

This pre-processing results in seven federated datasets $\mathcal{D}_{\text{ICU}}$, one per ICU, as listed in Table~\ref{table:baelines}. 
To support subsequent model development and evaluation, each ICU further splits its local dataset $\mathcal{D}_{\text{ICU}}$ into a training set $\mathcal{D}_{\text{ICU}}^{\text{Train}}$ (80\%) and a test set $\mathcal{D}_{\text{ICU}}^{\text{Test}}$ (20\%).
Among other things, Figure~\ref{fig:rnn} provides a conceptual illustration of the data structure used for each client in the federated setup.

\subsection{Sepsis Onset Prediction Model}

In this subsection, we describe the architecture of our sepsis onset prediction model, the organization of data into time-series windows for training, and the configuration used to enable FL across participating ICUs.

\noindent{\textbf{Data Windowing.}} To enable the prediction of sepsis onset, patient data is structured into multiple overlapping 6-hour input windows. This design choice is motivated by prior research demonstrating that predictions on patients' outcomes typically require at least 4 hours of observational data to achieve reliable performance \cite{pattalung2021comparison}. Additionally, input windows of 5 to 6 hours have been shown to be particularly effective for early sepsis detection tasks \cite{mondrejevski2023predicting}.

Following this approach, the initial window spans from the time of admission ($t=0$) to hour 5 ($t=5$), with the corresponding label indicating whether the patient develops sepsis by hour 30 ($t=30$). This yields a prediction window of 25 hours. A sliding-window approach is then applied, where each subsequent window is shifted forward by one hour. For example, the next sample includes data from $t=1$ to $t=6$ and aims to predict sepsis occurrence within the subsequent 24 hours. This process continues until the prediction window reduces to one hour, resulting in a dataset with varying prediction horizons.

\begin{figure*}[t]
    \centering
    \includegraphics[width=\textwidth]{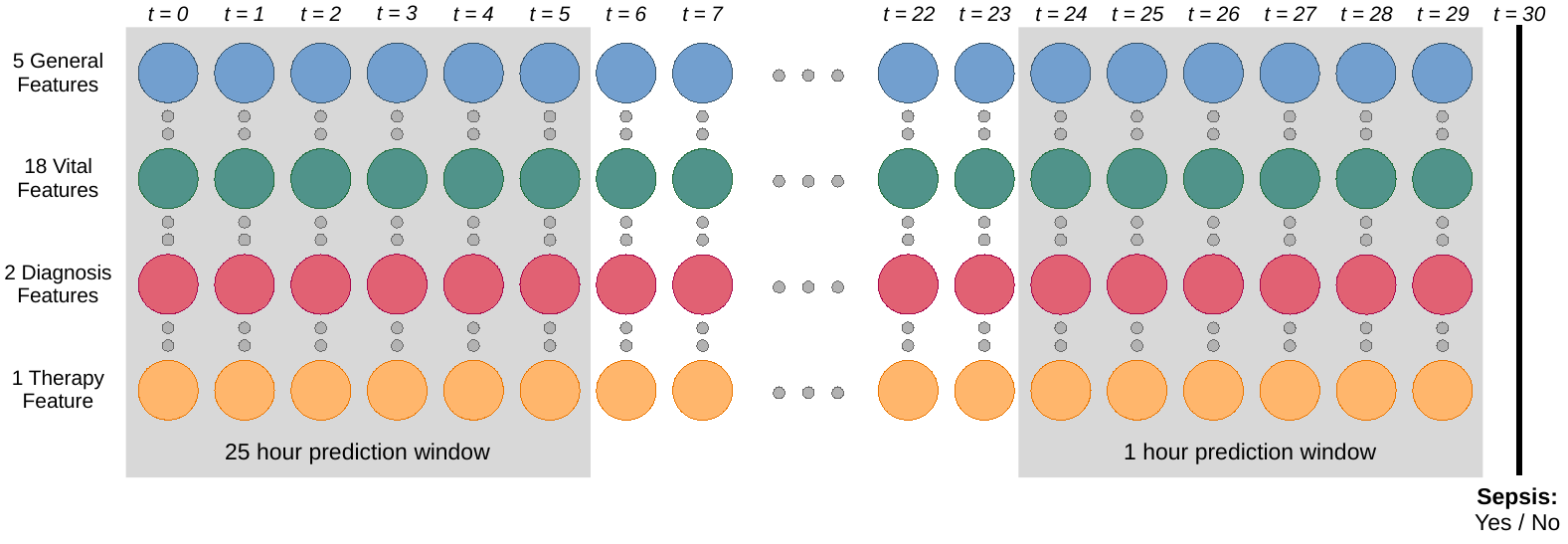}
    \caption{Time-Series Data and Sliding-Window Approach}
    \label{fig:rnn}
\end{figure*}

Figure~\ref{fig:rnn} illustrates this sliding-window mechanism and its alignment with the binary sepsis labels. This multi-horizon setup allows the model to learn from a spectrum of early-warning intervals, enhancing its ability to detect patterns indicating future sepsis onset. As a result, the model gains greater generalizability and is better suited for real-time clinical deployment, where predictions must adapt dynamically to the evolving state of a patient's condition.

\noindent{\textbf{Prediction Model.}} To leverage the temporal structure of the data and due to its proven effectiveness in clinical prediction tasks, we follow previous work and deploy a deep LSTM-based model for our sepsis onset prediction.

Our model builds on architectures proposed in prior FL studies (e.g., \cite{mondrejevski2022flicu}) and is enhanced with an attention mechanism \cite{mondrejevski2024masicu}, which allows the model to assign higher weights to clinically relevant time steps, capturing long-range dependencies. More precisely, the model's input layer is followed by a self-attention layer, then three LSTM layers with 16 units each. Each LSTM layer includes batch normalization and a dropout rate of 0.2 to prevent overfitting. A subsequent linear layer reduces the dimensionality to 8 units, again followed by batch normalization and dropout. Finally, a single linear output neuron provides a binary prediction indicating the likelihood of sepsis onset.

\noindent{\textbf{Federated Training.}} To train this model across institutions without centralizing data, we rely on the \textit{Federated Averaging} (FedAvg) algorithm \cite{mcmahan2017communication}. FedAvg is the most widely used aggregation strategy in FL, particularly in healthcare applications \cite{dusing2024integrating}. It aggregates local model updates from each hospital by computing a weighted average of their parameters, where weights correspond to the size of each hospital's local dataset \cite{kairouz2021advances}.

The federated training proceeds as follows: the central server initializes the global model and sends it to all participating ICUs. Each ICU then trains the model locally on its own data for three epochs and returns the updated model to the server. After receiving updates from all ICUs, the server aggregates them to form a new global model. This process is repeated for a total of 50 rounds, enabling the model to learn across all ICUs while preserving data privacy.

\subsection{Model Evaluation}

Model evaluation is conducted on two levels: an ICU-specific evaluation and an overall aggregated evaluation. The ICU-specific evaluation utilizes each ICU’s held-out test set $\mathcal{D}_{\text{ICU}}^{\text{Test}}$ to assess the extent to which each hospital benefits from the FL approach, particularly in terms of early and accurate sepsis onset prediction. The overall evaluation, by contrast, aggregates all clients’ test sets into a unified dataset $\mathcal{D}^{\text{Test}}$ and serves to quantify the global predictive capabilities of the model across institutions.

To contextualize the effectiveness of the proposed FL model, we compare three distinct evaluation settings during our analyses, namely \textit{Local}, \textit{Federated}, and \textit{Central}:
\begin{itemize}
    \item \textbf{Local:} The proposed LSTM model is trained independently at each ICU using only its local training set $\mathcal{D}_{\text{ICU}}^{\text{Train}}$.
    \item \textbf{Federated:} The proposed FL-based model is trained collaboratively across hospitals without sharing raw data, as described in the previous subsection.
    \item \textbf{Central:} The model is trained centrally on the aggregated training data $\mathcal{D}^{\text{Train}}$ from all ICUs, serving as a theoretical gold standard. Much like FL, it leverages all available data, but unlike FL, it does so without requiring frequent model aggregation, thus providing an upper bound on achievable performance in the absence of data privacy constraints.
\end{itemize}

To compare the performance across these evaluation settings, we utilize four different metrics: the \textit{F1-Score}, the \textit{Area Under the Curve} ($AUC$), and two use-case-specific metrics designed to capture clinical advantages in FL settings, namely the \textit{Federated Improvement Ratio} ($FIR$) and the \textit{Early Detection Advantage} ($EDA$). The former two metrics are standard and can be applied independently to each of the three evaluation settings. In contrast, $FIR$ and $EDA$ are designed specifically for the use-case at hand to compare the federated and local settings in order to quantify potential improvements gained through FL. These metrics are defined and motivated as follows:

\begin{itemize}
    \item \textbf{F1-Score:} The F1-Score balances the trade-off between precision and recall, providing a single value that captures both aspects of prediction quality. It is computed as:
    {\small\begin{equation}
        \text{F1-Score} = \frac{2 \cdot TP}{2 \cdot TP + FP + FN},
    \end{equation}}
    where $TP$, $FP$, and $FN$ denote \textit{True Positives}, \textit{False Positives}, and \textit{False Negatives}, respectively. This metric is especially useful as both false alarms and missed detections carry significant consequences in the clinical setting.
    \item \textbf{AUC:} $AUC$ measures the model's ability to distinguish between sepsis and non-sepsis cases across all classification thresholds. It is widely used in healthcare due to its robustness to threshold selection and its interpretability in terms of discriminative ability. It is defined as:
    {\small\begin{equation}
        AUC = \int_0^1 TPR(FPR^{-1}(x)) dx,
    \end{equation}}
    where $TPR$ and $FPR$ denote the \textit{True Positive Rate} and \textit{False Positive Rate}, respectively. We follow common implementations and rely on the trapezoidal rule to compute a practical approximation of the $AUC$ from discrete data points. Higher $AUC$ values indicate better overall performance across varying decision boundaries.
    \item \textbf{FIR:} $FIR$ is intended to quantify the improvement achieved by the federated model over the local baseline by comparing two sets: (1) sepsis cases correctly predicted by the federated model but missed by the local one, and (2) those correctly predicted by the local model but missed by the federated one. It is defined as:
    {\small\begin{equation}
        FIR = \frac{TP_{\text{Federated}} \cap FN_{\text{Local}}}{TP_{\text{Local}} \cap FN_{\text{Federated}}}. \label{eq:fir}
    \end{equation}}
    A ratio greater than 1 indicates that the federated model captures more true sepsis cases than it misses in comparison to the local baseline, highlighting a net diagnostic gain through the use of FL.
    \item \textbf{EDA:} $EDA$ captures the temporal benefit of the federated model by computing, for each correctly identified sepsis case, how many hours earlier the prediction was made compared to the local model. It is defined as:
    {\small\begin{equation}
        EDA = \frac{1}{|\mathcal{D}^{\text{Test}}_{\text{Sepsis}}|} \textstyle \sum_{d \in \mathcal{D}^{\text{Test}}_{\text{Sepsis}}} \left( t_{\text{Local}}^{(d)} - t_{\text{Federated}}^{(d)} \right), \label{eq:eda}
    \end{equation}}
    where $\mathcal{D}^{\text{Test}}_{\text{Sepsis}}$ is the subset of test data with sepsis onset, and $t_{\text{Local}}^{(d)}$, $t_{\text{Federated}}^{(d)}$ denote the earliest correct prediction time of sepsis onset by the local and federated models for case $d$, respectively. A positive $EDA$ indicates that the federated model recognizes sepsis onset earlier.
\end{itemize}

All evaluations are performed using 5-fold cross-validation, balancing reliability and computational cost. For each metric, we report the mean and standard deviation across folds to reflect both average performance and variability.

\section{Demonstration and Evaluation}

During model evaluation, we focus on three aspects: (1) the performance of the FL model compared to both local and central baselines, (2) a deeper temporal analysis of sepsis prediction quality, and (3) a comparison with fixed-prediction-window models in terms of performance and overhead.

\subsection{Federated Model Performance Comparison}

\noindent{\textbf{Baseline Comparison.}} We begin our evaluation by comparing the federated model to two baselines: a model trained locally at each ICU and a centralized data model. Table~\ref{table:baelines} shows the performance in terms of $F1$-Score and $AUC$ across all seven ICUs and the overall setting. The federated model consistently outperforms the local baselines across all ICUs. In some cases, it even surpasses the centralized model, though these differences are not statistically significant. Gains in $F1$-Score over local models range from modest to substantial, particularly in clients with more limited data. These results highlight the benefit of collaborative training across institutions, where FL compensates for local data scarcity or bias. Importantly, the federated model achieves an overall $AUC = 0.8057$, exceeding the clinically relevant threshold of 0.8 commonly cited for sepsis prediction models~\cite{ccorbaciouglu2023receiver}.

\begin{table}[t]
    \begin{center}
    \caption{Performance Comparison of FL With Local Baselines and Central Gold Standard (Improvements Over Local Baselines Bold and Underlined)}
    \label{table:baelines}
    \scalebox{0.86}{
        \begin{tabularx}{\textwidth}{c|c||Y|Y|Y|Y|Y|Y|Y||Y}
            & Model & MICU/ & \multirow{2}{*}{TSICU} & \multirow{2}{*}{CVICU} & \multirow{2}{*}{MICU} & \multirow{2}{*}{NSICU} & \multirow{2}{*}{SICU} & \multirow{2}{*}{CCU} & \multirow{2}{*}{Overall} \\
            & Training & SICU & &  &  &  &  &  & \\
            \cline{2-10}
            & Number of & \multirow{2}{*}{5,499} & \multirow{2}{*}{6,056} & \multirow{2}{*}{3,460} & \multirow{2}{*}{4,379} & \multirow{2}{*}{3,507} & \multirow{2}{*}{3,998} & \multirow{2}{*}{1,711} & \multirow{2}{*}{28,610} \\
            & Patients & & & & & & & & \\
            \hline\hline
            \multirow{6}{*}{\rotatebox[origin=c]{90}{$F1\text{-}Score$}}&\multirow{2}{*}{Local}& 0.7261 & 0.7680 & 0.8167 & 0.7574 & 0.5890 & 0.7251 & 0.7551 & 0.7522 \\
            & & \tiny{±0.01} & \tiny{±0.01} & \tiny{±0.04} & \tiny{±0.01} & \tiny{±0.05} & \tiny{±0.02} & \tiny{±0.01} & \tiny{±0.03} \\
            \cline{2-10}
            &\multirow{2}{*}{Federated}& \textbf{\underline{0.7791}}& \textbf{\underline{0.7953}} & \textbf{\underline{0.8482}} & \textbf{\underline{0.7811}} & \textbf{\underline{0.6389}} & \textbf{\underline{0.7803}} & \textbf{\underline{0.7745}}  & \textbf{\underline{0.8002}}\\
            & & \tiny{±0.02} & \tiny{±0.02} & \tiny{±0.01} & \tiny{±0.01} & \tiny{±0.04} & \tiny{±0.01} & \tiny{±0.01} & \tiny{±0.03} \\
            \cline{2-10}
            &\multirow{2}{*}{Central}& 0.7741 & 0.7876 & 0.8793 & 0.7816 & 0.7056 & 0.7439 & 0.7904  & 0.8127\\
            & & \tiny{±0.01} & \tiny{±0.02} & \tiny{±0.02} & \tiny{±0.02} & \tiny{±0.03} & \tiny{±0.01} & \tiny{±0.01} & \tiny{±0.03}\\
            \hline
            \hline
            \multirow{6}{*}{\rotatebox[origin=c]{90}{$AUC$}}&\multirow{2}{*}{Local}& 0.7221 & 0.7706 & 0.8334 & 0.7855 & 0.6136 & 0.7315 & 0.7568 & 0.7613\\
            & & \tiny{±0.01} & \tiny{±0.01} & \tiny{±0.04} & \tiny{±0.02} & \tiny{±0.04} & \tiny{±0.02} & \tiny{±0.02} & \tiny{±0.03}\\
            \cline{2-10}
            &\multirow{2}{*}{Federated}& \textbf{\underline{0.7907}} & \textbf{\underline{0.7799}} & \textbf{\underline{0.8478}} & \textbf{\underline{0.8087}} & \textbf{\underline{0.6340}} & \textbf{\underline{0.7689}} & \textbf{\underline{0.7925}}  & \textbf{\underline{0.8057}}\\
            & & \tiny{±0.01} & \tiny{±0.02} & \tiny{±0.03} & \tiny{±0.01} & \tiny{±0.04} & \tiny{±0.02} & \tiny{±0.01} & \tiny{±0.04}\\
            \cline{2-10}
            &\multirow{2}{*}{Central}& 0.7779 & 0.7926 & 0.8830 & 0.7813 & 0.7179 & 0.7699 & 0.7967 & 0.8195\\
            & & \tiny{±0.01} & \tiny{±0.01} & \tiny{±0.02} & \tiny{±0.02} & \tiny{±0.03} & \tiny{±0.02} & \tiny{±0.01} & \tiny{±0.03}\\
        \end{tabularx}
    }
    \end{center}
\end{table}

Given the strong correlation between $F1$-Score and $AUC$ observed in Table~\ref{table:baelines}, subsequent analyses report only $F1$-Scores for clarity.
\begin{figure*}[t]
    \centering
    \begin{subfigure}[b]{0.45\textwidth}
        \centering
        \includegraphics[width=1\textwidth]{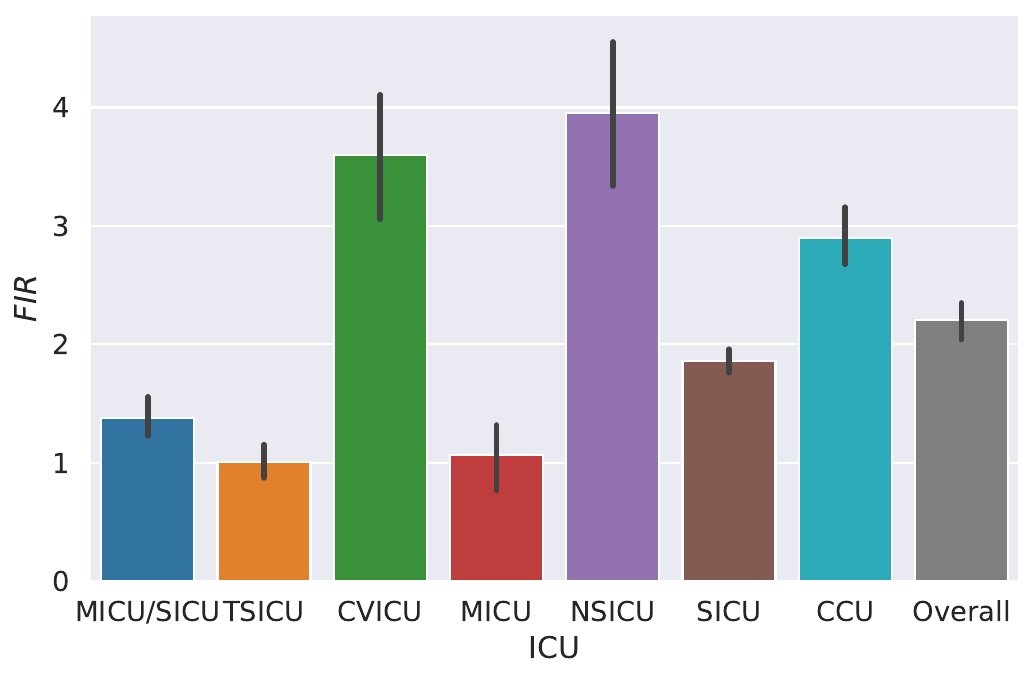}
        \caption{Federated Improvement Ratio}
    \end{subfigure}
    \begin{subfigure}[b]{0.45\textwidth}
        \centering
        \includegraphics[width=1\textwidth]{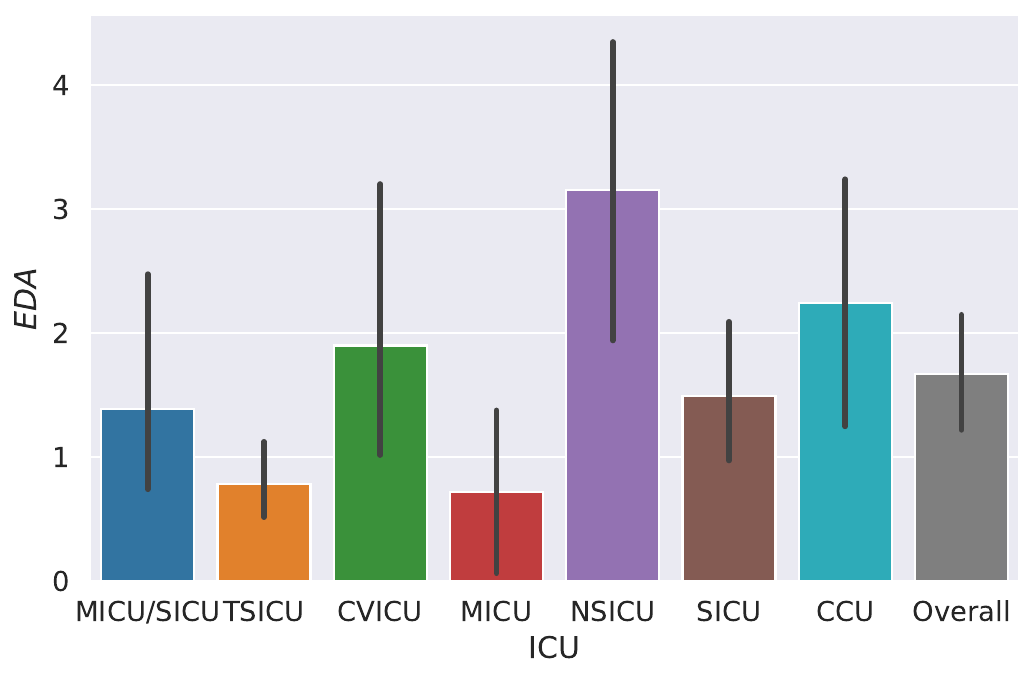}
        \caption{Early Detection Advantage}
    \end{subfigure}
    \caption{ICU-wise Comparison of Improvements Through FL}
    \label{fig:fir_eda}
\end{figure*}

\noindent{\textbf{Federated Improvement Analysis.}} To further quantify the benefit of FL, Figure~\ref{fig:fir_eda} (a) shows the $FIR$, which captures the relative change in correct sepsis detection compared to the local model. It shows that all ICUs except TSICU exhibit $FIR > 1$, indicating that the federated model identifies more sepsis cases than it misses relative to local models. CVICU, NSICU, and CCU achieve particularly strong gains with $FIR$ values above 3, whereas ICUs such as MICU/SICU show modest improvements. On average, the federated model correctly identifies twice as many sepsis cases missed by local models as vice versa.

\noindent{\textbf{Early Detection Advantage.}} To complement our performance comparison, we assess whether the federated model enables earlier sepsis detection. Figure~\ref{fig:fir_eda} (b) plots the average number of hours sepsis is detected earlier using the FL model compared to local baselines. All clients exhibit $EDA > 0$, confirming the benefit of FL in accelerating detection. The average improvement is close to 2 hours. Notably, NSICU shows the largest lead time gain (over 3 hours), followed by CVICU and CCU (around 2 hours). The remaining ICUs experience gains between 0.7 and 1.5 hours.

Overall, the federated model improves both detection accuracy and timeliness compared to local models and achieves performance close to a centralized gold standard -- all while preserving data privacy. This confirms the viability of FL for clinical prediction tasks across diverse hospital environments.

\subsection{Temporal Evaluation of Prediction Performance}

\begin{figure*}[t]
    \centering
    \begin{subfigure}[b]{0.45\textwidth}
        \centering
        \includegraphics[width=1\textwidth]{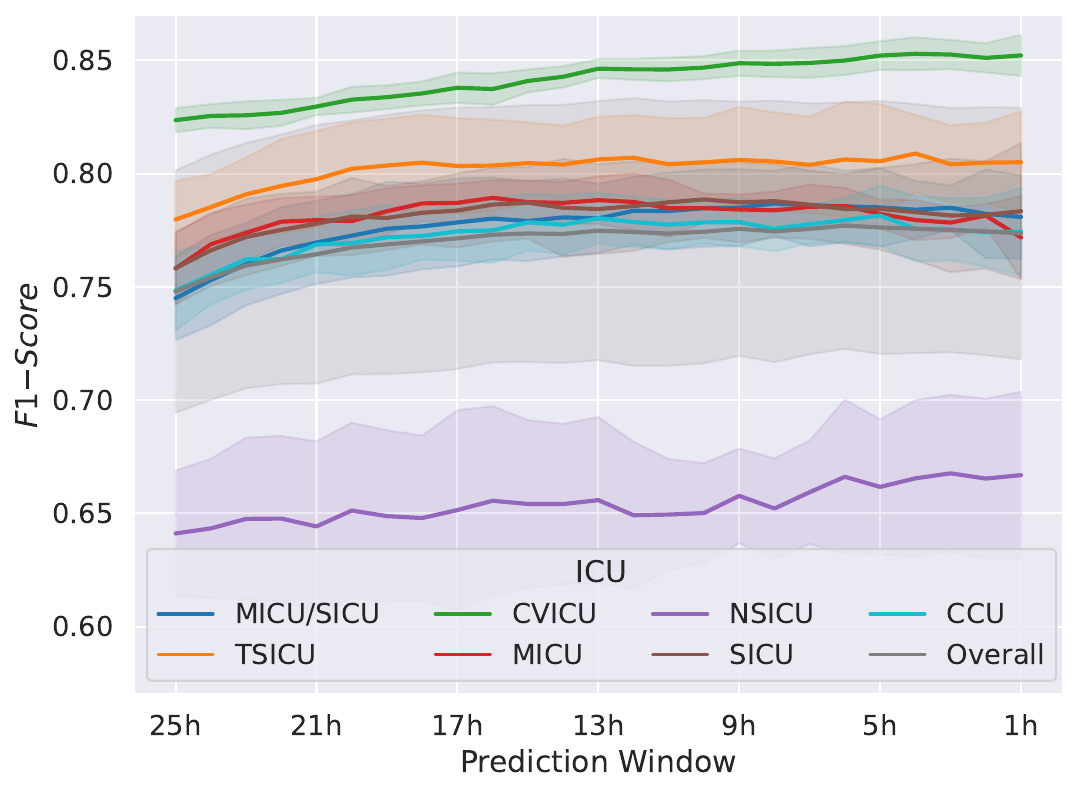}
        \caption{Federated $F1\text{-}Score$}
    \end{subfigure}
    \begin{subfigure}[b]{0.45\textwidth}
        \centering
        \includegraphics[width=1\textwidth]{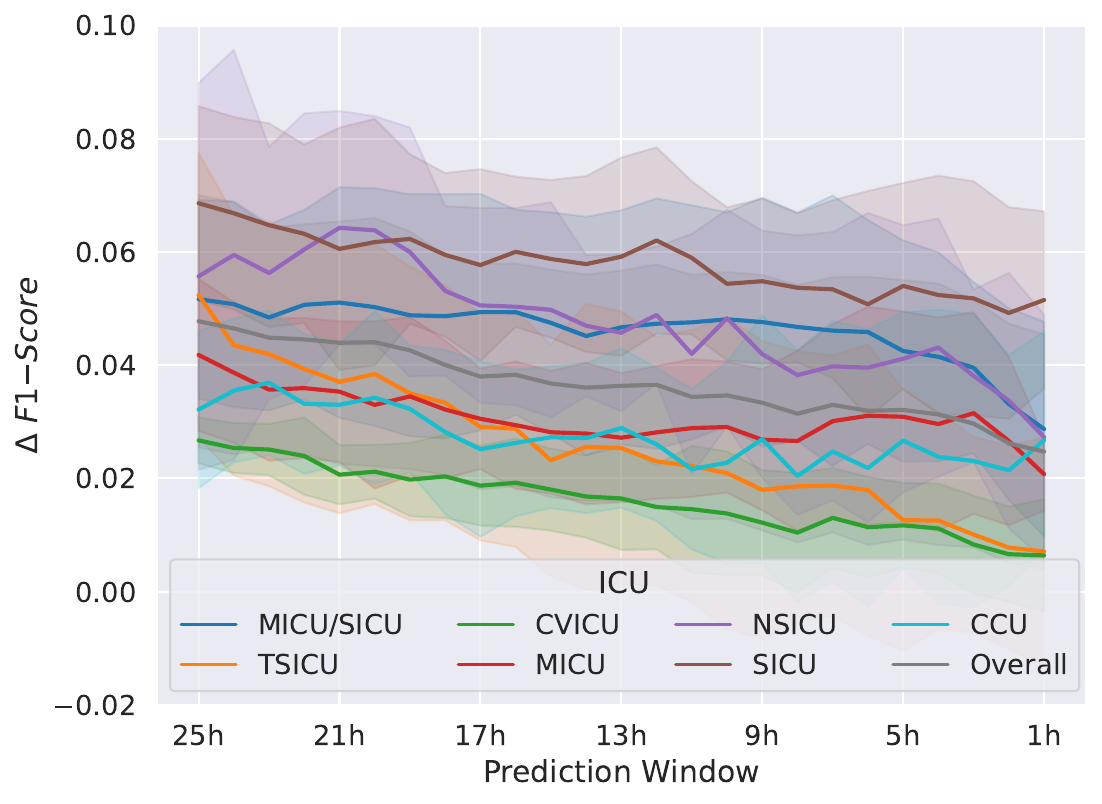}
        \caption{Improvements Over Local Baselines}
    \end{subfigure}
    \caption{Performance and Improvements Over Local Baselines Over Time}
    \label{fig:time}
\end{figure*}

\noindent{\textbf{Predictive Performance.}} Figure \ref{fig:time} (a) plots the F1-Score for each client for the different prediction windows (from 25h to 1h). In line with existing studies on the subject \cite{mondrejevski2023predicting,goh2021artificial,rosnati2021mgp,scherpf2019predicting}, we find that the prediction performance is better for short-term predictions, whereas long-term predictions are less accurate.

More specifically, we find that the performance for prediction windows smaller than 12h are mostly stable, where only few ICUs (e.g., CVICU and NSICU) improve when smaller prediction windows. For larger prediction windows, we observe a clear trend among all clients, where longer prediction horizons correlate with weaker performance in terms of F1-Score.
For most ICUs as well as the overall performance, the F1-Score starts at about 0.75 for 25h predictions and increases to about 0.79 for predictions below 12h horizons. 

Notable exceptions are CVICU, which has a significantly higher F1-Score starting at 0.82 and improving up to 0.85. Contrary, NSICU has significantly smaller scores, ranging from 0.64 to 0.67. These discrepancies are likely due to the fact that their data is dissimilar from the the data held by the remaining ICUs, which turns out favorable for CVICU and unfavorable of NSICU.

\noindent{\textbf{Improvements over Baseline.}} While the previous finding on the temporal aspects of predictive performance mostly confirmed existing studies, Figure \ref{fig:time} (b) plots the improvements achieved through the use of FL over its local baselines. This is a novel perspective on the evaluation of models for sepsis onset prediction.

The results show an almost linear trend among most ICUs as well as overall. The plot shows that the improvements over the local baselines are the largest for large prediction windows and decrease for smaller prediction horizons. This is interesting as our previous analysis proved weaker performance for such prediction windows, indicating that the local models suffer even more significantly from long prediction windows. 
The plot shows that the improvements for 25h predictions range between 0.02 and 0.07. For short term predictions such as 1h, improvements decrease to about 0.01 to 0.06. 
Notably, the CVICU client, which we previously found to perform best among all prediction windows, offers the smallest improvements over the local baselines, indicating a particularly strong local baseline model. These findings are also supported by the previous Table \ref{table:baelines}, where CVICU has, in fact, the highest baseline F1-Score.

The combined findings from Figure \ref{fig:time} show that while FL is also prone to deteriorating model performance for long-term predictions, we provide evidence that models trained federately are more resilient to these effects, indicated by the larger improvements over baselines for long-term predictions.

\subsection{Comparison with Fixed Prediction Window}

In earlier sections, we discussed the conceptual advantages of using a variable prediction window -- namely, reduced computational and organizational overhead, as well as greater clinical flexibility. Here, we complement those arguments by empirically comparing our proposed model against dedicated models trained for fixed prediction windows in terms of performance and computational cost.

\noindent{\textbf{Predictive Performance.}} 
Table~\ref{table:fixed} compares the F1-Score of our variable-prediction-window model to three representative fixed-prediction-window models trained independently for the exemplary prediction windows of 25h, 15h, and 5h. While the fixed-window models show slightly higher scores in several cases, most differences fall within one standard deviation and are therefore not statistically significant. Notable exceptions include MICU/SICU and SICU, where two of the three fixed-window models significantly outperform the variable-window model. In contrast, for NSICU, our variable-prediction-window model achieves better performance in two of the three comparisons.

\begin{table}[t]
    \begin{center}
    \caption{F1-Score Improvements of Fixed-Prediction-Windows Over Variable-Prediction-Windows (Bold and Underlined if Exceeding Standard Deviation)}
    \label{table:fixed}
    \scalebox{0.86}{
        \begin{tabularx}{\textwidth}{c||Y|Y|Y|Y|Y|Y|Y||Y}
            Prediction & MICU/ & \multirow{2}{*}{TSICU} & \multirow{2}{*}{CVICU} & \multirow{2}{*}{MICU} & \multirow{2}{*}{NSICU} & \multirow{2}{*}{SICU} & \multirow{2}{*}{CCU} & \multirow{2}{*}{Overall} \\
            Window & SICU &  &  &  &  &  &  & \\
            \hline\hline
            \multirow{2}{*}{25h}& \textbf{\underline{0.0265}} & 0.0281 & 0.0091 & 0.0265 & -0.0037 & \textbf{\underline{0.0277}} & 0.0116 & 0.0203 \\
            & \tiny{±0.02} & \tiny{±0.03} & \tiny{±0.03} & \tiny{±0.04} & \tiny{±0.05} & \tiny{±0.02} & \tiny{±0.02} & \tiny{±0.03} \\
            \cline{1-9}
            \multirow{2}{*}{15h}& \textbf{\underline{0.0286}} & 0.0268 & -0.0119 & 0.0201 & 0.0059 & 0.0193 & 0.0160 & 0.0171 \\
            & \tiny{±0.02} & \tiny{±0.03} & \tiny{±0.02} & \tiny{±0.02} & \tiny{±0.05} & \tiny{±0.02} & \tiny{±0.03} & \tiny{±0.02} \\
            \cline{1-9}
            \multirow{2}{*}{5h}& 0.0293 & 0.0297 & 0.0115 & \textbf{\underline{0.0275}} & -0.0159 & \textbf{\underline{0.0246}} & 0.0192 & \textbf{\underline{0.0202}} \\
            & \tiny{±0.03} & \tiny{±0.03} & \tiny{±0.02} & \tiny{±0.02} & \tiny{±0.06} & \tiny{±0.02} & \tiny{±0.03} & \tiny{±0.02}\\
        \end{tabularx}
    }
    \end{center}
\end{table}

\noindent{\textbf{Computational Overhead.}} 
To quantify computational cost, we measured training time under identical conditions using the same hardware (NVIDIA Tesla V100 16GB). Average training time per round was comparable across settings -- 109 seconds for our variable-window model versus a combined 115 seconds for the all fixed-window models. This is expected since data size and model complexity are constant, and aggregation overhead is minimal. However, convergence behavior differs significantly. The variable-prediction-window model converged within an average of 5.3 rounds, while the fixed-prediction-window models required approximately 15.4 rounds on average -- nearly three times as many. This difference is likely caused by the fact that each fixed-prediction-window model is trained on a smaller data subset, limiting generalization and convergence.

The previous comparison does not even account for some variable factors affecting the total overhead. For example, training multiple models instead of a single one using FL increases the communication overhead involved in the training, which scales linearly with the number of models. 
Furthermore, it disregards the substantial organizational overhead associated with maintaining and validating multiple specialized models in practice.

Taken together, these results suggest that our variable prediction window approach offers a substantially more efficient and scalable alternative, with only minimal, often statistically insignificant trade-offs in predictive performance. We argue that the computational and operational benefits far outweigh the marginal performance gains occasionally observed with dedicated fixed-window models.

\section{Conclusion, Limitations, and Future Work}

In this study, we present a FL approach for early sepsis onset prediction based on an attention-enhanced LSTM architecture. Our model is specifically designed to support variable prediction windows, allowing it to operate flexibly across a range of clinical scenarios, ranging from immediate risk assessment at ICU admission to long-term forecasts. Compared to existing approaches that rely on fixed prediction horizons \cite{mondrejevski2023predicting,goh2021artificial}, our model generalizes across multiple prediction windows in a single training process, reducing computational overhead without significant performance trade-offs.

Through extensive evaluation on the MIMIC-IV dataset, we demonstrate that the federated model consistently outperforms local baselines across all participating ICUs and performs competitively with a centrally trained upper-bound model. These findings validate that FL not only enables collaborative model development across institutions but also improves early detection of sepsis -- a critical requirement for clinical decision support. In particular, we show that FL provides the greatest performance improvements in long-term predictions, where local models deteriorate in performance. This contributes a novel and clinically meaningful insight to the literature, addressing a gap in prior works that largely focused on short-term prediction accuracy.

\noindent{\textbf{Limitations.}} To start with the obvious, our experiments are conducted on a single dataset and task only. Although the MIMIC-IV dataset is well-established in the field of medical research, this limits the generalizability of our findings. Moreover, our FL setup assumes ideal conditions, ignoring practical aspects of FL application, such as issues with client dropout, client synchronization, and increased deployment complexity. Additionally, we did not include baselines other than the local and central training of the same LSTM model we train federately.

\noindent{\textbf{Future Work.}} In the future, we aim to focus on validating our approach in clinical settings  and using external datasets to assess generalizability in real-world settings. We also plan to integrate techniques facilitating model explainability and transparency -- an essential factor for clinical adoption. Lastly, we aim to explore and mitigate potential failure modes in FL settings, including issues arising from data imbalance, malicious participants, or unreliable clients, to further strengthen robustness and practical viability of our approach.

\end{document}